\documentclass[letterpaper, 10 pt, conference]{ieeeconf}  

\IEEEoverridecommandlockouts                              

\usepackage{graphicx} 
\usepackage{amsmath} 
\usepackage{amssymb}  
\usepackage{amsfonts}
\usepackage{xcolor}
\usepackage{epsfig}
\usepackage{graphicx}
\usepackage{subfigure}
\usepackage{epstopdf} 

\usepackage{comment}
\usepackage{graphicx}

\usepackage{algorithm}
\usepackage{algpseudocode}
\usepackage{bm}
\usepackage{bbm}
\usepackage{multirow}

\usepackage{csquotes}

\title{\LARGE \bf
Embracing Safe Contacts with Contact-aware Planning and Control
}

\author{Zhaoting Li $^{1}$, Miguel Zamora $^{1}$, Hehui Zheng $^{1,2}$, Stelian Coros $^{1}$ 
\thanks{$^{1}$ Computational Robotics Lab, Computer Science Department, ETH, Zurich, Switzerland. %
{\tt\small zhaotli@student.ethz.ch \{miguel.zamora, stelian.coros\}@inf.ethz.ch}}
\thanks{$^{2}$ Soft Robotics Lab, Department of Mechanical and Process Engineering, ETH, Zurich, Switzerland. %
{\tt\small hehui.zheng@srl.ethz.ch} 
}%
}

\begin{document}

\maketitle
\thispagestyle{empty}
\pagestyle{empty}

\begin{abstract}


Unlike human beings that can employ the entire surface of their limbs as a means to establish contact with their environment, robots are typically programmed to interact with their environments via their end-effectors, in a collision-free fashion, to avoid damaging their environment. 
In a departure from such a traditional approach, this work presents a contact-aware controller for reference tracking that maintains interaction forces on the surface of the robot below a safety threshold in the presence of both rigid and soft contacts. Furthermore, we leveraged the proposed controller to extend the BiTRRT sample-based planning method to be contact-aware, using a simplified contact model.
The effectiveness of our framework is demonstrated in hardware experiments using a Franka robot in a setup inspired by the Amazon stowing task. A demo video of our results can be seen here: https://youtu.be/2WeYytauhNg


\end{abstract}

\section{INTRODUCTION}

\begin{figure}[t]
	\centering
	\includegraphics[scale=0.26]{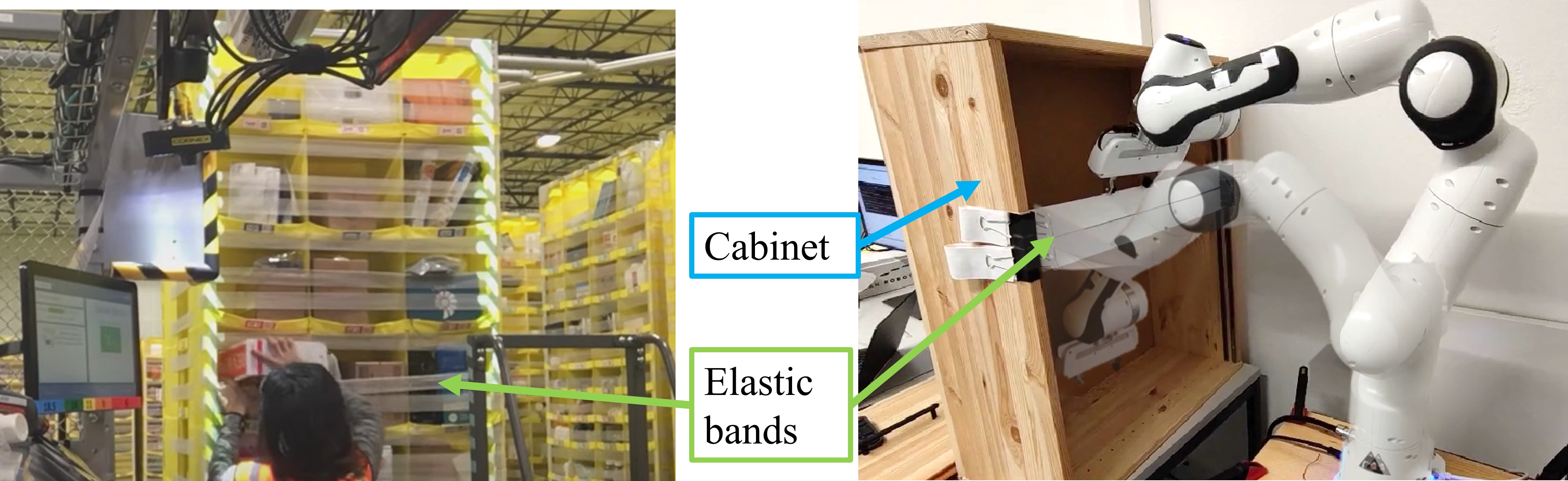}
	\caption{Left: the Amazon stowing task. Right: The test environment in this work. During the execution of the goal-reaching task, the robot needs to embrace two types of contacts: contact with elastic bands, which results in their deformation so that the robot can reach the goal state, and possible contact with the cabinet, arising from inaccuracies of the cabinet's position.}
	\label{fig:introduction_env}
\end{figure}

Contact-rich tasks require robots to make contact with their environment. A good example of such tasks is the stowing task in the Amazon warehouses, where elastic bands are mounted on cabinets to prevent objects from falling out, and human operators establish contact with the elastic bands using their entire arms to make room to insert new items in the cabinets. This work focuses on  goal-reaching tasks in a similar environment, as shown in Fig. \ref{fig:introduction_env}. 
To enable robots to perform this task, safe contact methods need to be developed to ensure secure interactions with the elastic bands.

Collision avoidance is a common method to ensure the robot's safety, but it may not be effective in uncertain environments due to perception errors. 
Additionally, in cluttered environments, finding a feasible solution that satisfies collision avoidance constraints can be challenging. 
Contact-aware/driven controllers \cite{2022_contact_aware_control_pang}, \cite{2019_contactDriven_posture_behavior}, \cite{2013_tactile_reaching_clutter}, \cite{2021skin_nullspace} have been proposed to address these issues, 
utilizing the robot's null space to reach a goal state while maintaining contact forces below a safety threshold.
However, these methods assume either known stiffness parameters for soft contact or known rigid contact, lacking a general algorithm for both types. 
Additionally, local minima can be a problem when using contact-aware controllers, and global planning algorithms are necessary. 
Due to the possible multi-modal characteristics of the planning problem, the receding horizon planning may also be stuck into the local minimum \cite{2013_tactile_reaching_clutter} \cite{2022allowingsafecontact_planning}.




To solve the aforementioned limitations, we propose a contact-aware controller that unifies rigid and soft contacts by estimating the stiffness parameters online. 

We derive a simple yet effective quasi-static dynamics model for soft contacts, which generalizes to the rigid contact dynamics proposed in \cite{2022_contact_aware_control_pang}. Furthermore, we introduce a contact-aware planning method that finds near-optimal trajectories, minimizing the deformation of the environment and preventing our controller from being stuck in local minima, when reaching into the cabinet environment in Fig. \ref{fig:introduction_env}.
The contributions of our work are the following:

\begin{itemize}
    \item A general contact-aware controller framework that unifies both the soft and rigid contacts.
    \item A global contact-aware planning algorithm that minimizes the deformation of the elastic band in the stowing task while reaching a goal state. 
\end{itemize}

\section{Methods: Contact-aware control}
\label{section::Contact-aware control}

\subsection{Preliminaries}

\subsubsection{Joint impedance controller}

Simple impedance controllers can be modeled as follows \cite{2013_tactile_reaching_clutter}:
The input of the controller $\bm q_{cmd}$ is called the 'virtual trajectory', as shown in Fig. \ref{fig:contact_notation}(a).
The control law is:
\begin{equation}
\label{eq:impendace_controller}
    \bm \tau = \bm K_q (\bm q_{cmd} - \bm q) - \bm D_q \dot{\bm q} + 
\hat{ \bm g}(\bm q).
\end{equation}

\subsubsection{Contact notation}
\label{section::contact_notation}
\begin{figure}[h]
  \centering
  \includegraphics[width=0.5\textwidth]{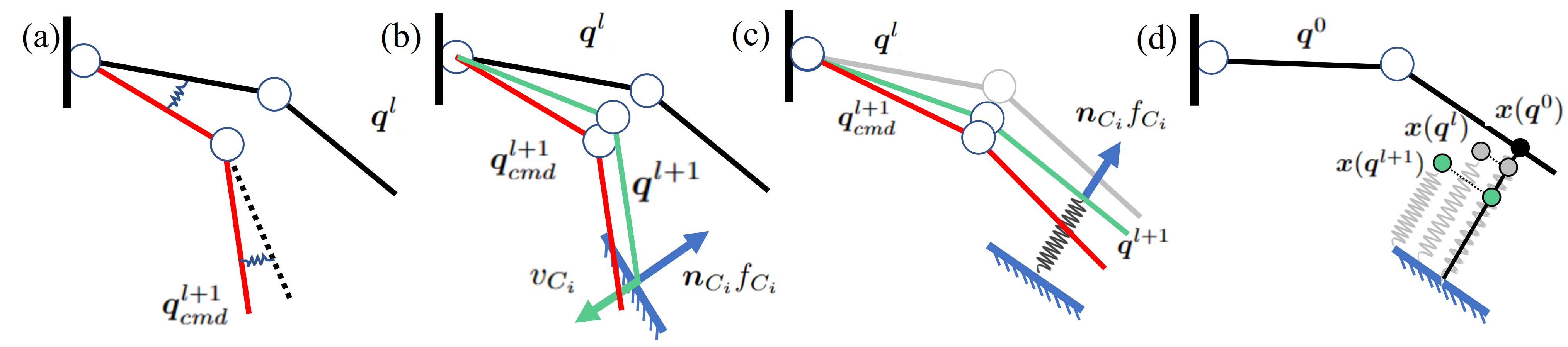}
  \caption{(a) An illustration of the joint impedance controller. (b) A contact blocks the robot when it follows the $\bm q_{cmd}^{l+1}$ command. (c) The robot has soft contact with the environment when following joint commands. (d) The robot's movement causes deformation in the linear spring model.}
  \label{fig:contact_notation}
\end{figure}

Denote the number of external contacts as $n_c$.
The position Jacobian at the $i$-th contact point $\bm p_{C_i}$ 
is $\bm J_{C_i} (\bm q, \bm p_{C_i})$.
The contact force can be expressed as $ \bm f_{C_i} = \bm n_{C_i} f_{C_i}$,
where the $\bm n_{C_i}$ is the unit vector representing the contact direction and $ f_{C_i}$ is its norm, as shown in Fig. \ref{fig:contact_notation}(b). 
Following \cite{2019_contactDriven_posture_behavior}, the reduced contact Jacobian is defined as 
$\bm J_{u_i} = \bm n_{C_i}^T \bm J_{C_i} (\bm q, \bm p_{C_i})$,
which relates the joint velocity to the linear contact point velocity along the contact direction:
\begin{align*}
v_{C_i} &= \bm n_{C_i}^T {\bm v}_{p_{C_i}}  = n_{C_i}^T  \bm J_{C_i} (\bm q, \bm p_{C_i}) \dot{\bm q}  =  \bm J_{u_i}  \dot{\bm q}.
\end{align*}
Then the external torque is:
$\bm \tau_{{ext}} = \sum_{i=1}^{n_C}\bm \tau_{{ext}_i} = \bm J_u^T \bm f_c$,
where $
    \bm{J}_u^T = [ \bm{J}_{u_1}^T  \cdots \bm{J}_{u_{n_C}}^T],
 \bm f_c = [
 f_{C_0}  \cdots  f_{C_{n_C}}]^T$.

\subsubsection{Quasi-static dynamics for hard contacts}

The quasi-static dynamics of hard contact can be expressed as an optimization problem \cite{2022_contact_aware_control_pang}, which minimizes the robot's potential energy subject to the constraint of zero contact velocity:
\begin{align}
 \mathop{\min}_{\bm q^{l+1}} \frac{1}{2} \| \bm q_{cmd}^{l+1} - \bm q^{l+1} \|^2_{\bm K_q} \text{ s.t. }
 \bm J_u(\bm q^l) (\bm q^{l+1} - \bm q^l) = 0,
\end{align}
where the index $l$ denotes the time stamp. 
The KKT optimality condition of the QP is:
\begin{subequations}
\begin{align}
     \bm K_q (\bm q^{l+1} - \bm q_{cmd}^{l+1} ) - \bm J_u^T
     \bm f_c
     =& \bm 0  \label{eq::hardcontraint01} \\
     \bm J_u (\bm q^{l+1} - \bm q^l) =& 0
\end{align}
\end{subequations}

\subsection{Problem formulation}

In this work, we consider two tasks for the robot to execute: 
(1) reaching a goal state $\bm q_{goal}$,
(2) ensuring that the contact force does not exceed a predefined safety threshold.

\subsubsection{Null space as backbone}
Denote the task velocity vector $\dot{\bm x}_i \in \mathbf{R}^{m_i \times 1}, i=1, \dots, r$, $ m_i \leq n_q$.
 As proposed in \cite{1991_nullspace_joint_velocity_solution}, a generic $i$-th task can be characterised by the differential kinematic equation
 $ \dot{\bm x}_i = \bm J_i(\bm q) \dot{\bm q}$.
Since the primary task Jacobian $\bm J_1$ has dimension $m_1 < n_q$, there is a kinematic redundancy of $n_q - m_1$ to accomplish low-priority tasks in the null space $\bm N_1$ of  $\bm J_1$ using $\dot{\bm q}_0$.
\begin{align}
\label{eq::nullspace_framework}
    \dot{\bm q} = \bm J_1^{\#} \dot{\bm x}_1 + \bm N_1 \dot{\bm q}_0, 
\end{align}
where $\bm N_1 = \bm I - \bm J_1^{\#} \bm J_1$.
It can be shown that Eq. \ref{eq::nullspace_framework} is also the solution to the following QP problem
\begin{align}
\label{eq::nullspace_qp}
    \mathop{\min}_{\dot{\bm q}}  \frac{1}{2}   \| \dot{\bm q} - \dot{\bm q}_0 \|^2 \quad
    \text{s.t. } \bm J_1(\bm q) \dot{\bm q} = \dot{\bm x}_1. 
\end{align}
Hence, $\dot{\bm q}_0$ in Eq. \ref{eq::nullspace_framework} can be viewed as the desired joint velocity given the constraint of satisfying the first task.
To describe the force-regulating task, the control law can be defined by $\bm J_1(\bm q) \dot{ \bm q} = \bm f_c - \bm f_d$, where $\bm f_d \in [\bm f_{min}, \bm f_{max}]$ denotes the desired force value. 

\begin{figure}[th]
  \centering
  \includegraphics[width=0.4\textwidth]{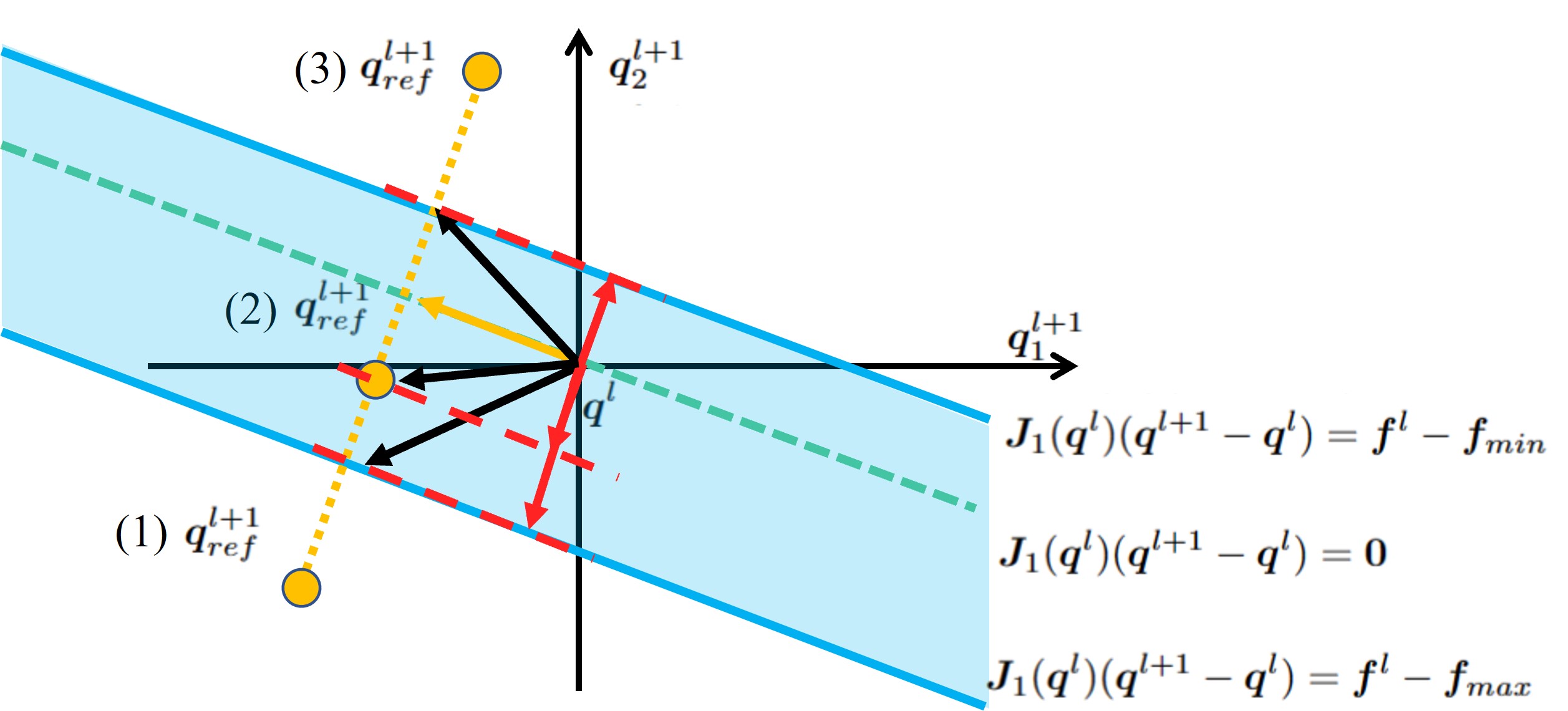}
  \caption{Three cases of explicitly determining $\bm f_d$. The blue lines denote the force-regulating task with $\bm f_d \in [\bm f_{min}, \bm f_{max}]$. The red arrow denotes the particular solution of the force-regulating task, and the yellow arrow denotes the null space correction. The black arrow denotes the final solution. 
  }
  \label{fig:nullspace_f_desired_choice}
\end{figure}

The QP formulation defined in Eq. \ref{eq::nullspace_qp} can then be modified for the contact-aware control task
\begin{align}
\label{eq::nullspace_force_task}
    \mathop{\min}_{\bm q ^{l+1}}  \frac{1}{2}   \|\bm q ^{l+1} - \bm q ^{l+1}_{ref} \|^2 
    \text{s.t. }  \bm J_1 (\bm q ^{l+1} - \bm q ^{l}) = \bm f^l - \bm f_d,
\end{align}
where the value of the desired force $\bm f_d$ should be determined explicitly.
Possible situations are shown in Fig. \ref{fig:nullspace_f_desired_choice}.
However, determining an accurate value for $\bm f_d$ is challenging in practice, and assigning a wrong value to it can lead to dangerous behaviors.

\subsubsection{Inequality constraint formulation}
\label{section::general_Inequality constraint formulation}
Inequality constraints can be utilized to keep the contact force inside a safety interval.
The general framework is defined as follows:
\begin{subequations}
\label{eq::general_optimization_framework}
\begin{align}
    \mathop{\min}_{\bm q ^{l+1}}  \frac{1}{2}   \|\bm q ^{l+1} - \bm q ^{l+1}_{ref} \|^2 \quad
    \text{s.t. } &\\ 
    \bm f^{l + 1}, \bm q^{l + 1} = \bm h( \bm q^{l+1}_{ref}, \bm q^l, \bm f^l) &\\
    \bm f_{min} \leq \bm f^{l+1} \leq \bm f_{max} &
\end{align}
\end{subequations}
where $\bm h(\cdot, \cdot, \cdot)$ is a model of the contact dynamics.
Note that the only difference w.r.t the null space framework is the use of a different method to describe the force-regulating task.


\subsection{Contact-aware control for unknown stiffness contacts}
\subsubsection{Quasi-static dynamics for soft contacts}

In this section, we derive the quasi-static dynamics for the soft contacts, which follow the contact model defined in Eq. \ref{eq::general_optimization_framework}.
Let $\bm q^0$ be the robot state in contact with the environment with zero deformation. 
The position of the contact point is denoted by $\bm x(\cdot): \bm q \rightarrow \mathbb{R}^3$. 
We also make the assumption that the frictional component can be neglected and the contact can be modeled as the linear elastic spring model \cite{2013_tactile_reaching_clutter}, \cite{2021skin_nullspace}:
\begin{equation}
\label{eq::spring_force_model}
   \bm f_c^{l+1} - \bm f_c^{l} =  
   \bm K_C \bm n_C^T( \bm x(\bm q^{l+1}) - \bm x(\bm q^{l})).
\end{equation}

As shown in Fig. \ref{fig:contact_notation}(d), given the current state $\bm q^l$ and the next command state $ \bm q_{cmd}^{l+1}$, the next state $\bm q^{l+1}$ can be obtained by minimizing the total energy of the whole system:
\begin{equation}
 \mathop{\min}_{\bm q^{l+1}} \frac{1}{2} \| \bm q_{cmd}^{l+1} - \bm q^{l+1} \|^2_{\bm K_q}
 + \frac{1}{2}  \| \bm n_C^T (\bm x(\bm q^{l+1}) - \bm x(\bm q^0) )\|^2_{\bm K_C}
\end{equation}
For the unconstrained quadratic optimization problem, the optimal solution can be found by letting the gradient be zero.
The optimality condition of this unconstrained QP is: 
\begin{subequations}
\label{eq::softconstraint}
\begin{align}
\bm K_q ( \bm q^{l+1} - \bm q_{cmd}^{l+1}) - \bm J_u ^T \bm f_c^{l+1} =
\bm 0 & \label{eq::softconstraint01}\\
\bm f_{c}^{l+1} -\bm f_{c}^{l} = -\bm K_{c} \bm J_{u} (\bm q^{l+1} - \bm q^l). \label{eq::softconstraint02}
\end{align}
\end{subequations}
 Eq. \ref{eq::softconstraint01} is equivalent to Eq. \ref{eq::hardcontraint01} in the optimality condition of the hard contact model, indicating that the soft contact and hard contact formulations are highly related. 
As $\bm K_c \rightarrow + \infty$ naturally enforces the constraint that $\bm J_{u} (\bm q^{l+1} - \bm q^l) \rightarrow \bm 0$, which is the zero-velocity constraint of the hard contact case.


\subsubsection{Online identification of the contact stiffness}
When dealing with uncertain environments, such as those involving both soft and rigid contacts, it's necessary to estimate the value of the stiffness matrix $\bm K_c$. 
The contact stiffness can be identified using the recursive least squares method (RLS). 
The contact model in Eq. \ref{eq::softconstraint02} can be seen as a linear system.
$ \bm z(k) = \bm H(k) \bm \phi + \bm w(k),a$
where $\bm z(k) =\bm f_{c}^{k+1} -\bm f_{c}^{k} $, $\bm \phi = -\bm K_{c}$,
$\bm H(k) \in \mathbb{R}^{n_c \times n_c}$.
We set the estimator's initial state  $\hat{\bm \phi }(0) = + \infty, \bm P(0) =\bm P_{\bm \phi } = \text{Var}[\bm \phi ]$.
The estimated stiffness average and variance are updated recursively using the force measurement from a contact detection algorithm. 

\subsubsection{Contact-aware control}

This section formulates the contact-aware control for both soft and rigid contacts by combining the general framework in Eq. \ref{eq::general_optimization_framework} and the quasi-static dynamics of soft contacts:
\begin{subequations}
\label{eq::contact_aware_control_soft}
\begin{align}
 \mathop{\min}_{\bm q^{l+1}, \bm q^{l+1}_{cmd},\bm f_{c}^{l+1}}  \| \bm q_{cmd}^{l+1} - \bm q_{ref}^{l+1} \|^2 + \epsilon \|\bm q^{l+1}_{cmd} -\bm q^{l}_{cmd} \|^2  &  \label{eq::sub::contact_aware_control_soft_a} \\
 \text{ s.t. } \quad \quad \quad \bm K_q (\bm q^{l+1} - \bm q_{cmd}^{l+1} ) - \bm J_u^T \bm f_{c}^{l+1}  = \bm 0 & \label{eq::sub::contact_aware_control_soft_b}\\
   \bm f_{c}^{l+1}  -\bm f_{c}^{l} = -\bm K_{c} \bm J_{u} (\bm q^{l+1} - \bm q^l)&  \label{eq::sub::contact_aware_control_soft_c}\\
    \bm f_{c}^{l+1} \leq f_{max} \bm I &  \label{eq::sub::contact_aware_control_soft_d}\\
    | \bm q^{l+1}_{cmd} - \bm q^{l}_{cmd} | \leq \Delta \bm q_{max}  \label{eq::sub::contact_aware_control_soft_e}
\end{align}
\end{subequations}
This quadratic programming problem has two equality constraints (\ref{eq::sub::contact_aware_control_soft_b}) and (\ref{eq::sub::contact_aware_control_soft_c}) that describe the quasi-static dynamics of soft contact.
The force regulating task is then described by the inequality constraint (\ref{eq::sub::contact_aware_control_soft_d}), where the lower bound of the force is removed because the contact is uni-lateral.
The objective (\ref{eq::sub::contact_aware_control_soft_a}) is modified from Eq. \ref{eq::general_optimization_framework}, to mitigate the negative effects of neglecting frictional forces, according to \cite{2022_contact_aware_control_pang}.
Note that the objective of this QP can be easily changed to other tasks, as detailed in section \ref{section::general_Inequality constraint formulation}.



\section{Contact aware planning}





\subsection{Problem definition}

A robot is denoted as $\mathcal{R}$ and its configuration space as $\mathcal{C}_{\mathcal{R}}$. 
For a robot with $n_{q_r}$ joints, $\mathcal{C}_{\mathcal{R}}$ is defined as $\mathbb{R}^{n_{q_r}}$.
$\bm q_r \in \mathcal{C}_{\mathcal{R}}$ represents the robot's joint state in the configuration space. 
We restrict the contacts to only occur on the allowed links.
The robot's link set is denoted as $l_{\mathcal{R}} = \{ l_i, i = 1, \dots, n_l \}$, where $n_l$ is the number of robot's links. 
We use $l_i(\bm q_r)$ to denote the 3d space that the $i$-th link occupies in state $\bm q_r$. 
The allowed link set is defined as $l_{\mathcal{R}}^a \in  l_{\mathcal{R}}$ and can be specified via prior knowledge or by removing all links that may cause the robot blocked when interacting with the world.

\begin{figure}[ht]
	\centering
	\includegraphics[scale=0.25]{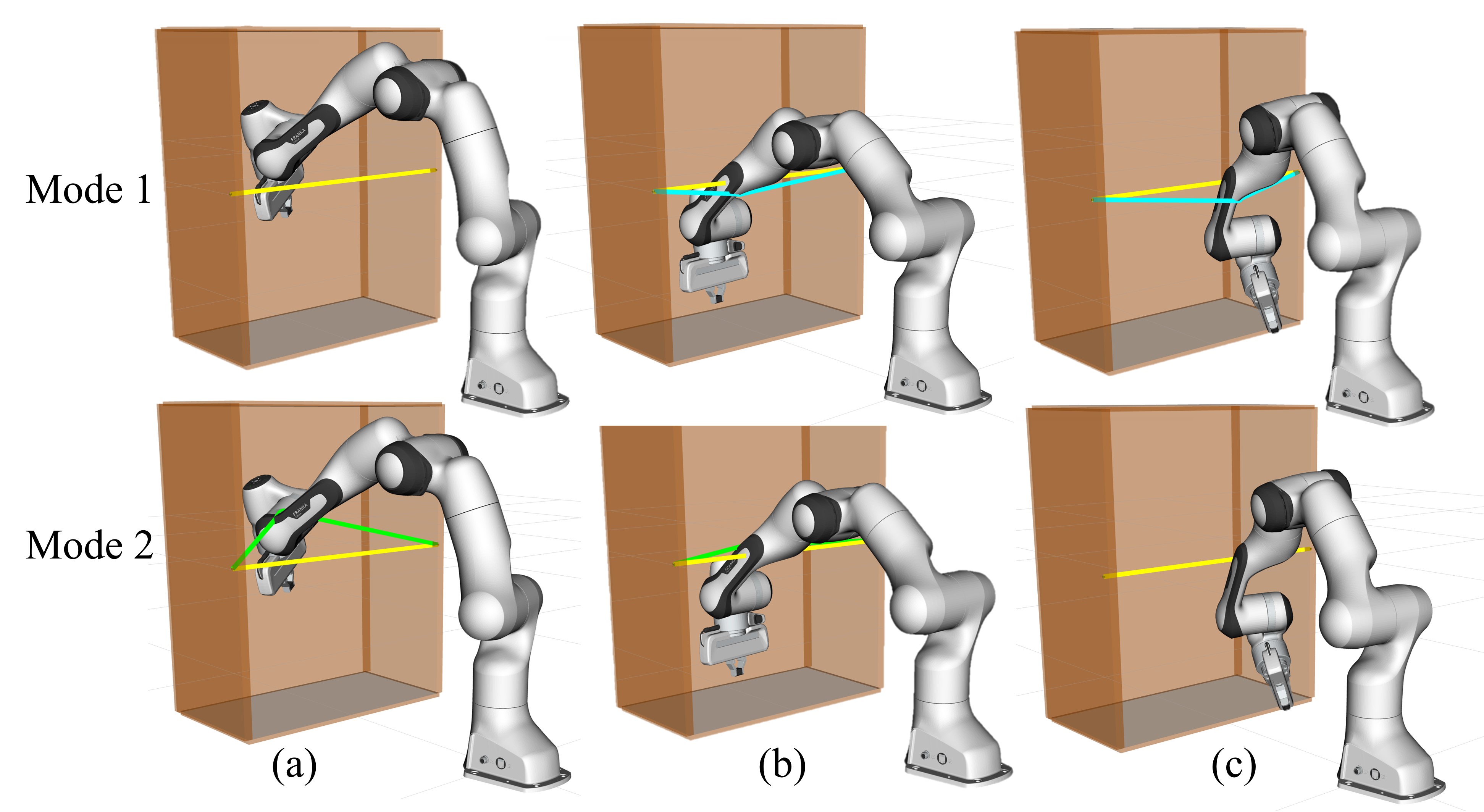}
	\caption{Examples of the elastic band configuration for different modes. The yellow line denotes the elastic band without deformation. The blue and green line denote the band deformed in the mode 1 and 2, respectively.}
	\label{fig:two_modes_simplified_eb}
\end{figure}

\subsubsection{The configuration space of the elastic band}
\label{section::simplified_model_of_eb}
\begin{figure}[ht]
	\centering
	\includegraphics[scale=0.175]{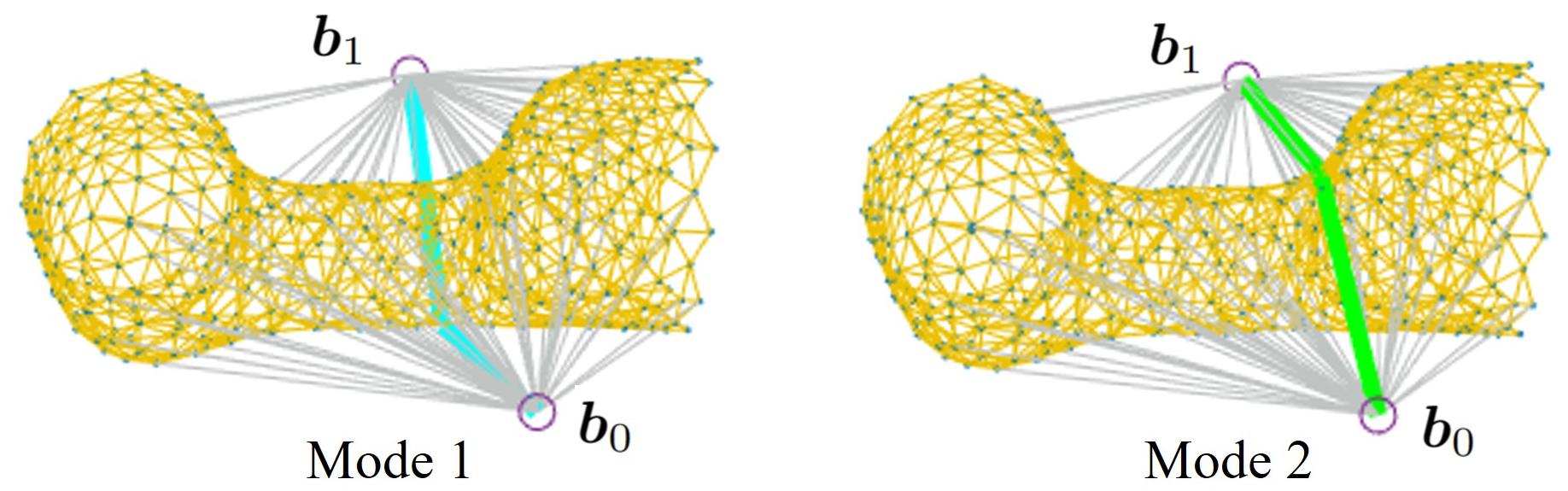}
	\caption{An example of the simplified model of the band. The graph of the robot surface is denoted by yellow lines. Gray lines denote the edges connecting $\bm b_0$ or $\bm b_1$ to their valid vertices. The obtained paths of the A* algorithm for modes 1 and 2 are visualized in blue and green curves, respectively.}
	\label{fig:simplified_model_eb}
\end{figure}

Let $\mathcal{B}$ be an elastic band with configuration space $\mathcal{C}_{\mathcal{B}}$. 
We assume that the start point $\bm b_0$ and the end point $\bm b_1$ of the band have the same height in the world frame.
Accordingly, we can define the interaction modes
 $ \sigma \in \mathcal{C}_{\sigma} = \{ 0, 1, 2 \}$.
Mode $\sigma = 0$ represents no contacts,
while $\sigma=1$ and $\sigma=2$ correspond to the robot being above and below the band, respectively. 
Examples of these modes are shown in Fig. \ref{fig:two_modes_simplified_eb}.
To simplify the interactions between elastic bands and robot, we assume that the direction of band deformation $\bm d_b$ is determined by the interaction mode $\sigma$ and the direction of the robot link in contact with the elastic band $\bm l_r$.
Under these assumptions, the elastic band can be simplified as a tuple
$(\bm b_0, \bm b_1, \bm d_b, L  )$,
where $\bm d_b = (-1)^{\mathbbm{1}_{\sigma}} (\bm b_0 - \bm b_1) \times \bm l_r $, and $\times$ denotes the cross product operation.
The length $L \in \mathcal{C}_L = [L_{min}, L_{max}] \subseteq \mathbb{R}$ denotes the shortest path between $\bm b_0$ and $\bm b_1$ without penetrating the robot surface, which can be obtained by the A* search \cite{Astar1968}, as shown in Fig. \ref{fig:simplified_model_eb}.

\subsubsection{The configuration space of the whole system}
The configuration space of the entire system is $\mathcal{C}=\mathcal{C}_{\mathcal{R}}\times\mathcal{C}_{\mathcal{B}}$, where the state of the system is denoted as $\bm q_s=[\bm q_r, \bm q_b]$, with $\bm q_b\in\mathcal{C}_{\mathcal{B}}$ representing the state of the elastic band.

To determine a valid configuration space of the whole system $\mathcal{C}^{valid}$, we first define $\mathcal{C}_{\mathcal{B}}^{valid}$ and $\mathcal{C}_{\mathcal{R}}^{valid}$.
The state $\bm q_b$ is valid if the elastic band is not under excessive tension beyond its elastic limit.
Define the valid configuration space as 
$
\mathcal{C}_{\mathcal{B}}^{valid} = \{ 
    \bm q_b | L \leq L_{max}
    \}$.
For the robot, the state $\bm q_r$ is valid if the robot has no collisions with other obstacles and only allowed links make contacts with the elastic band:
\begin{align}
    \mathcal{C}_{\mathcal{R}}^{valid} = \{ 
    \bm q_r | \neg \texttt{Collision}(\mathcal{O}, l_i(\bm q_r)),l_i \in l_{\mathcal{R}} \nonumber \\
    \cup \neg \texttt{Collision}( l_j(\bm q_r), \mathcal{B}), l_j \in l_{\mathcal{R}}  \setminus l_{\mathcal{R}}^a
    \}.
\end{align}
Then the valid configuration space of the whole system can be defined as $\mathcal{C}^{valid} = \mathcal{C}_{\mathcal{R}}^{valid} \times \mathcal{C}_{\mathcal{B}}^{valid}$.

The whole system is under-actuated because the state $\bm q_b$ of the elastic band cannot change without moving the robot.
Given the $\bm b_0$ and $\bm b_1$,
we assume that the $\bm q_b$ can be uniquely defined by the robot state $\bm q_r$ and the specified mode $\sigma$.
Therefore, we can define the model of the elastic band as 
$\bm q_b = f_b (\bm q_r, \sigma)$, where $f_b(\cdot, \cdot): \mathcal{C}_{\mathcal{R}} \times \mathcal{C}_{\sigma} \rightarrow \mathcal{C}_{\mathcal{B}}$.


The space of the goal state $\mathcal{Q}_s^{goal}$ is defined by the Cartesian task of the robot.
Denote the initial and goal state of the whole system as $
\bm q_s^{init}$ and $
\bm q_s^{goal} \in \mathcal{Q}_s^{goal}$.
The cost of the state $\bm q_s$ is defined as a function $c(\cdot): \mathcal{C} \rightarrow \mathbb{R}^{+}$.
Define a sequence of robot configurations, ${  \bm q_r^{init}, \bm q_r^{1}, \dots, \bm q_r^{n_{\pi}} }$, which can be linearly interpolated to obtain the path $\pi_r(\cdot): [0, n_{\pi}] \rightarrow \mathcal{C}_{\mathcal{R}}$ with $\pi_r(0) = \bm q_r^{init}$.
Formally, the contact-aware planning problem is formulated as follows:
\begin{subequations}
\label{eq::contact_aware_planning}
\begin{align}
   \mathop{\min}_{\sigma} \mathop{\min}_{\bm q_r^{1}, \dots, \bm q_r^{n_{\pi}}} \int_0^{n_{\pi}} c \big(\pi_r(s), f_b(\pi_r(s), \sigma) \big)  ds, \text{ s.t. }\\
   \pi_r = \texttt{Path}(\bm q_r^{init}, \bm q_r^{1}, \dots, \bm q_r^{n_{\pi}}) \label{eq::contact_aware_planning::c} \\
   \pi_s( n_{\pi}) \in \mathcal{Q}_s^{goal} (\sigma) \\
    \pi_s(s) \in \mathcal{C}^{valid}, \forall s \in [0, n_{\pi}]
\end{align}
\end{subequations}

\subsection{Contact-aware control to optimize the joint goal state}
\label{section::contact-aware-control_optimized joint goal state}

In the planning loop, instead of estimating and bounding the contact force, we can minimize the elastic band's deformation while satisfying the end-effector constraints. 
This approach utilizes the robot's null space to achieve this goal.
Mathematically, we want to find the optimal goal state $\bm q^{goal*}_s(\sigma) \in \mathcal{Q}_s^{goal} (\sigma)$ for a given mode $\sigma$. 
Note that reaching the end-effector position is treated as the first-priority task and used as the constraint of the QP formulation.
\begin{subequations}
\label{eq::optimizied_q_goal_contact_aware}
\begin{align}
    \mathop{\min}_{\bm q^{l+1}_r} \| \bm J_u (\bm q^{l+1}_r - \bm q^l_r) - \gamma  \Delta L_{band}  \|^2 \text{ s.t. } &\\
    \bm J_e  (\bm q^{l+1}_r - \bm q^l_r) = \bm 0 \\
     | \bm q^{l+1}_r - \bm q^{l}_r | \leq \Delta \bm q_{max} 
\end{align}
\end{subequations}
where $\bm J_e$ is the task Jacobian for the end-effector, and $\bm J_u$ and $ \Delta L_{band}$  are obtained by the simplified model of the elastic band.
We should also take into account the other obstacles in the environment. 
Hence, during each iteration of Eq. \ref{eq::optimizied_q_goal_contact_aware}, we need to check whether the robot collides with other obstacles.


\subsection{Contact aware planning to find feasible trajectory to the joint goal state}

This section considers the sub-problem of the contact-aware planning problem in Eq. \ref{eq::contact_aware_planning}.
For each mode $\sigma$, we reduce the search space by restricting the goal state to be the optimal goal state $\bm q^{goal*}_s(\sigma)$ obtained in section \ref{section::contact-aware-control_optimized joint goal state}.
Accordingly, the constraint in Eq. \ref{eq::contact_aware_planning::c} is replaced by $\pi_s( n_{\pi}) = \bm q^{goal*}_s (\sigma)$.

Our planner builds upon the BiTRRT algorithm (Bidirectional Transition-based RRT \cite{2010TRRT}, \cite{2013BiTRRT}) and incorporates modifications to consider potential interactions between the robot and elastic band.
The key variation is to sample random robot state $\bm q_r^{rand}$ instead of the whole configuration space $\mathcal{C}$.
Accordingly, we modify the \texttt{distance} function inside the \texttt{NearestNeighbor} function and the \texttt{attemptLink} function to account for the distance between $\bm q_b$.
Denote the  \texttt{distance} function for the configuration space $\mathcal{C}, \mathcal{C}_{\mathcal{R}}, \mathcal{C}_{\mathcal{B}}$ as $d_s(\cdot, \cdot), d_r(\cdot, \cdot), d_b(\cdot, \cdot)$, respectively, 
where $d_r(\bm q_{r1}, \bm q_{r2}) = \| \bm q_{r1} - \bm q_{r2} \|$,
\begin{equation*}
\label{eq::distance_eb}
    d_b(\bm q_{b1}, \bm q_{b2}) =  
   \left\{ {\begin{array}{*{30}{c}}
	{  \| L_1 - L_2 \|, \sigma_1 = \sigma_2 \text{ or } \sigma_1 = 0 \text{ or } \sigma_2 = 0}\\
        { +\infty, \text{others} }
	\end{array}} \right. 
\end{equation*}
then the distance in the whole configuration space $\mathcal{C}$ is defined as $d_s(\bm q_{s1}, \bm q_{s2})^2 = d_r(\bm q_{r1}, \bm q_{r2})^2 + \lambda_b  d_b(\bm q_{b1}, \bm q_{b2})^2$, where $\lambda_b > 1$ is the scaling factor to emphasize the distance in $\mathcal{C}_{\mathcal{B}}$. 
By considering the mode of $\mathcal{C}_{\mathcal{B}}$ in the Eq. \ref{eq::distance_eb},
it effectively prevents connecting configurations that violate the physics constraint by setting the distance between such configurations to infinity. 

\section{Experiments}

\begin{figure}[!h]
	\centering
	\includegraphics[scale=0.25]{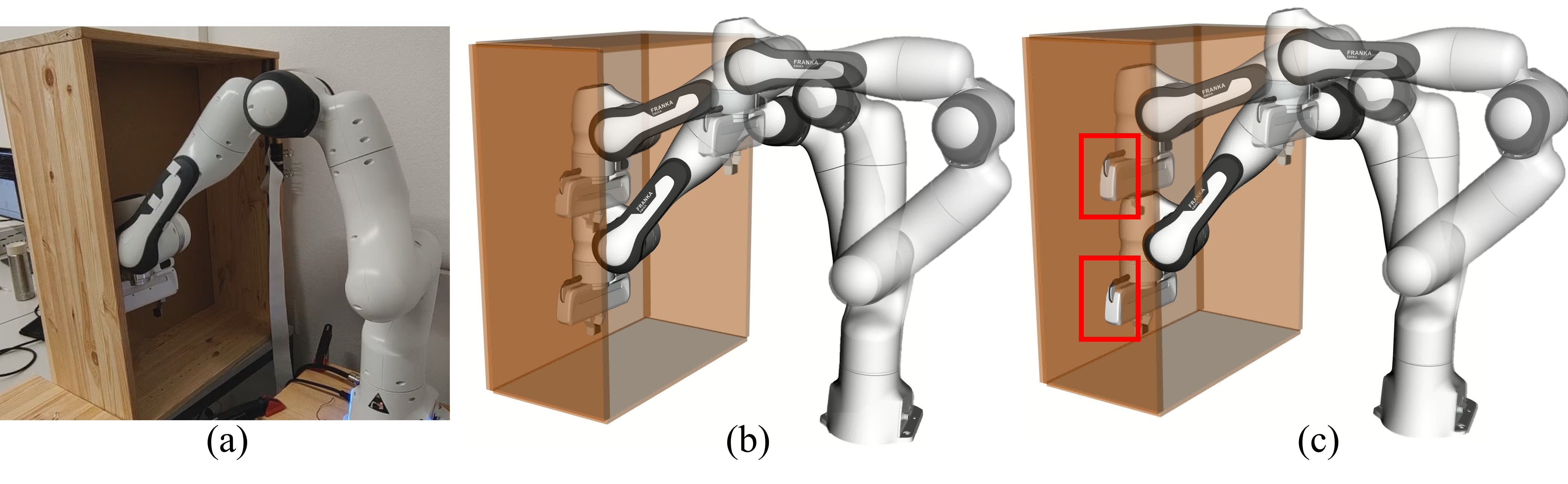}
	\caption{Rigid contact task: (a) the final robot state. (b): the recorded trajectory of the robot before reaching the state (a). (c): the reference trajectory of the robot, which leads to collision with the cabinet.}
	\label{fig:rigid_contact_trajectory}
\end{figure}

We tested the contact-aware control method using a Franka robot. 
For both the soft and the rigid contact environments (Fig. \ref{fig:rigid_contact_trajectory} and Fig. \ref{fig:introduction_env} respectively), our controller can keep the contact force below a predefined threshold while tracking the reference trajectory as closely as possible.

\begin{figure}[ht]
	\centering
	\includegraphics[scale=0.22]{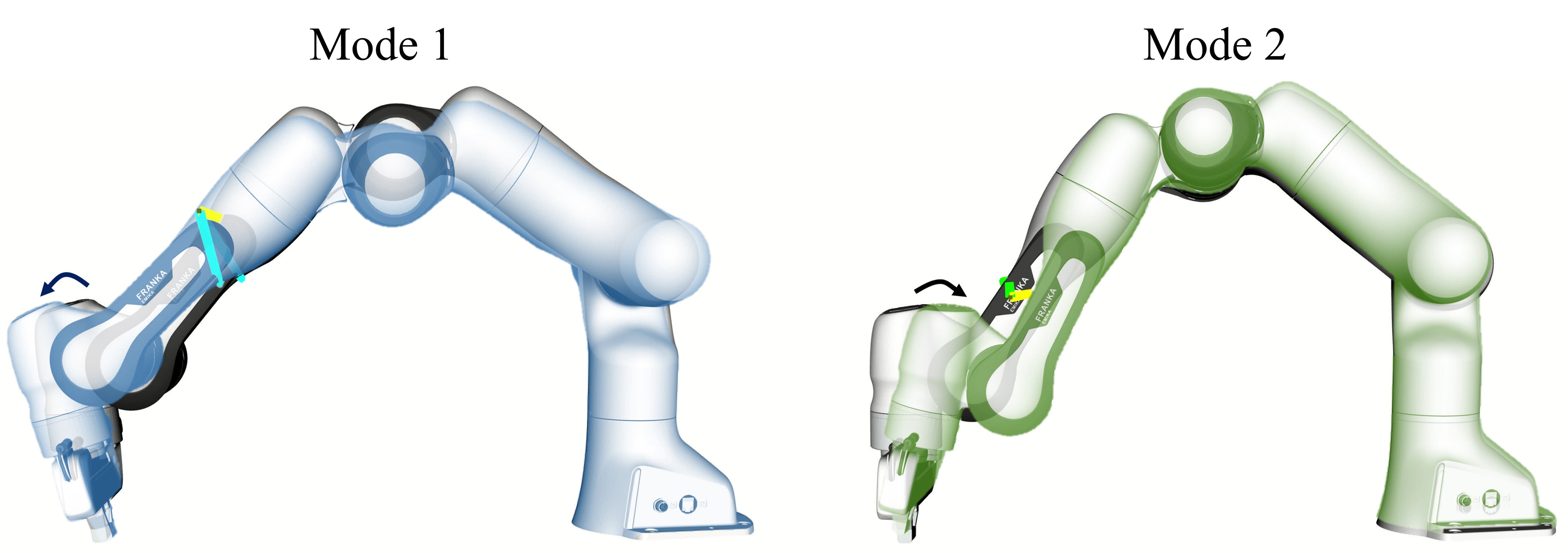}
	\caption{Experiment results of the optimization of the joint state goal. Robot with the normal color: the initial goal state obtained by the inverse kinematics. Robot visualized in blue and green: the optimized goal state that minimizes the deformation of the elastic band in modes 1 and 2, respectively.}
	\label{fig:optimized_joint_goal_state}
\end{figure}

\begin{figure}[ht]
	\centering
	\includegraphics[scale=0.2]{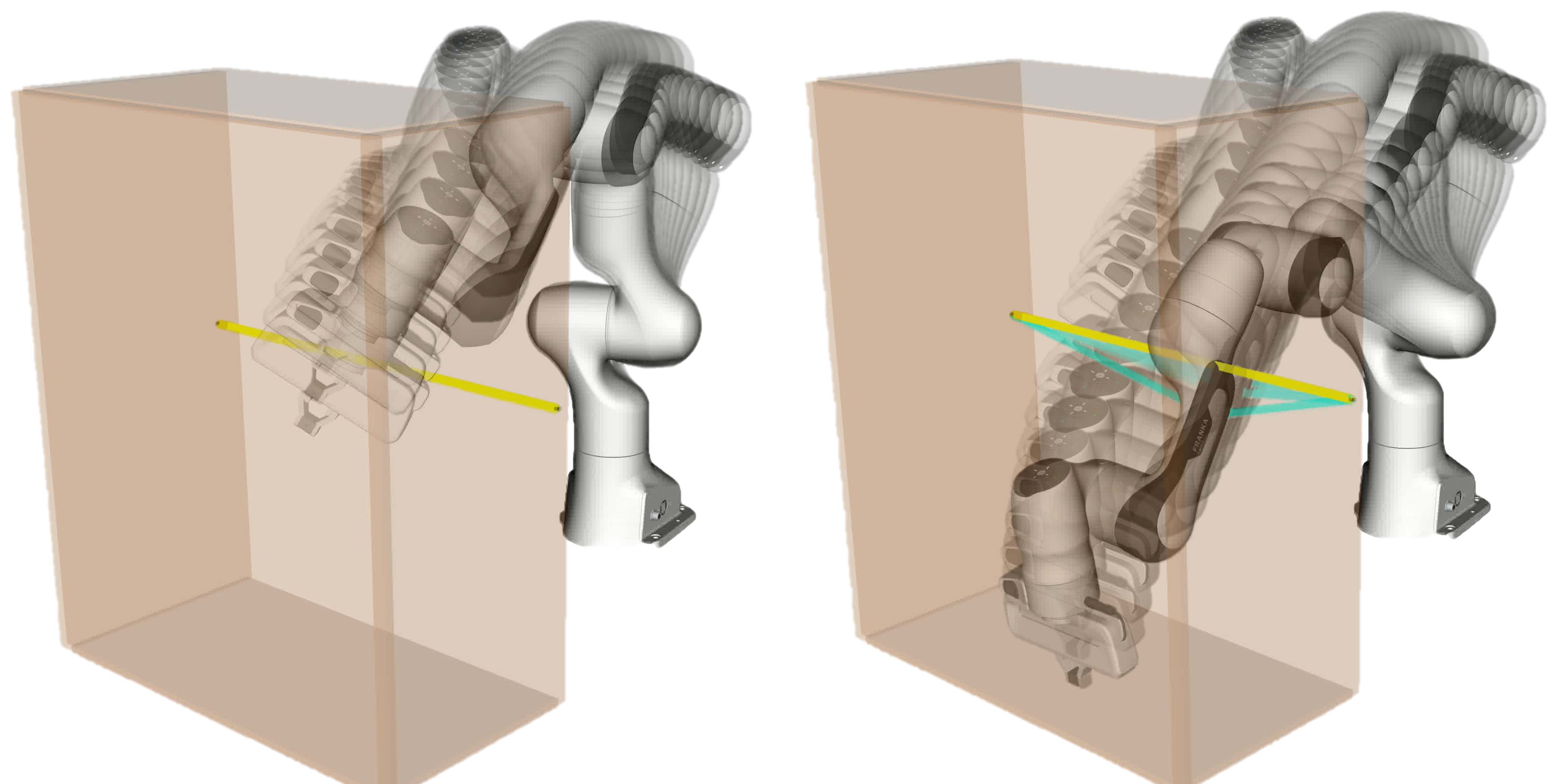}
	\caption{The result of the contact-aware planning under mode 1. }
	\label{fig:Planning_trajectory_display_mode0}
\end{figure}

Examples of the optimized joint goal states (section \ref{section::contact-aware-control_optimized joint goal state}) are shown in Fig. \ref{fig:optimized_joint_goal_state}, and a planned path for mode 1 is shown in Fig. \ref{fig:Planning_trajectory_display_mode0}.
The robot first raises its end-effector to keep it from being blocked by the elastic band. 
Then it reaches inside the cabinet while deforming the band.

\section{Conclusions and future work}
Our contact-aware control and planning methods can enable robots to make physical contact via their surface with the environment in a safe manner. Future work includes the extension of the planning algorithm to other types of environments, like cluttered environments, considering safety constraints for objects that might tip over.

\bibliographystyle{plain}
\bibliography{root}

\begin{thebibliography}{1}

\bibitem{2021skin_nullspace}
Alessandro Albini, Francesco Grella, Perla Maiolino, and Giorgio Cannata.
\newblock Exploiting distributed tactile sensors to drive a robot arm through
  obstacles.
\newblock {\em IEEE Robotics and Automation Letters}, 6(3):4361--4368, 2021.

\bibitem{2013BiTRRT}
Didier Devaurs, Thierry Sim{\'e}on, and Juan Cort{\'e}s.
\newblock Enhancing the transition-based rrt to deal with complex cost spaces.
\newblock In {\em 2013 IEEE international conference on robotics and
  automation}, pages 4120--4125. IEEE, 2013.

\bibitem{Astar1968}
Peter~E Hart, Nils~J Nilsson, and Bertram Raphael.
\newblock A formal basis for the heuristic determination of minimum cost paths.
\newblock {\em IEEE transactions on Systems Science and Cybernetics},
  4(2):100--107, 1968.

\bibitem{2010TRRT}
L{\'e}onard Jaillet, Juan Cort{\'e}s, and Thierry Sim{\'e}on.
\newblock Sampling-based path planning on configuration-space costmaps.
\newblock {\em IEEE Transactions on Robotics}, 26(4):635--646, 2010.

\bibitem{2013_tactile_reaching_clutter}
Advait Jain, Marc~D Killpack, Aaron Edsinger, and Charles~C Kemp.
\newblock Reaching in clutter with whole-arm tactile sensing.
\newblock {\em The International Journal of Robotics Research}, 32(4):458--482,
  2013.

\bibitem{2019_contactDriven_posture_behavior}
Mikael Jorda, Elena~Galbally Herrero, and Oussama Khatib.
\newblock Contact-driven posture behavior for safe and interactive robot
  operation.
\newblock In {\em 2019 International Conference on Robotics and Automation
  (ICRA)}, pages 9243--9249. IEEE, 2019.

\bibitem{2022_contact_aware_control_pang}
Tao Pang and Russ Tedrake.
\newblock Easing reliance on collision-free planning with contact-aware
  control.
\newblock In {\em 2022 International Conference on Robotics and Automation
  (ICRA)}, pages 8375--8381. IEEE, 2022.

\bibitem{1991_nullspace_joint_velocity_solution}
Siciliano~B Slotine and B~Siciliano.
\newblock A general framework for managing multiple tasks in highly redundant
  robotic systems.
\newblock In {\em proceeding of 5th International Conference on Advanced
  Robotics}, volume~2, pages 1211--1216, 1991.

\bibitem{2022allowingsafecontact_planning}
Xinghao Zhu, Wenzhao Lian, Bodi Yuan, C~Daniel Freeman, and Masayoshi Tomizuka.
\newblock Allowing safe contact in robotic goal-reaching: Planning and tracking
  in operational and null spaces.
\newblock {\em arXiv preprint arXiv:2211.08199}, 2022.

\end{thebibliography}



\end{document}